\pdfoutput=1

\documentclass[11pt]{article}

\usepackage{EMNLP2022}

\usepackage{times}
\usepackage{enumitem}
\usepackage{latexsym}
\usepackage{multirow}
\usepackage{multicol}
\usepackage{graphicx}
\usepackage{booktabs}
\usepackage{amsthm,amsmath,amssymb}
\usepackage[T1]{fontenc}

\usepackage[utf8]{inputenc}
\usepackage{CJKutf8}
\usepackage{microtype}

\usepackage{inconsolata}

\title{Cross-Align: Modeling Deep Cross-lingual Interactions for Word Alignment}

\author{Siyu Lai\textsuperscript{1}\thanks{ \ \ Work done when Siyu were interning at Pattern Recognition Center, WeChat AI, Tencent Inc, China.},
Zhen Yang\textsuperscript{2} , 
Fandong Meng\textsuperscript{2},
\textbf{Yufeng Chen}\textsuperscript{1}\thanks{ \ \ Yufeng Chen is the corresponding author.}, \\ 
\textbf{Jinan Xu}\textsuperscript{1} and \textbf{Jie Zhou}\textsuperscript{2}\\
\textsuperscript{1}Beijing Key Lab of Traffic Data Analysis and Mining, \\
Beijing Jiaotong University, Beijing, China \\
\textsuperscript{2}Pattern Recognition Center, WeChat AI, Tencent Inc, China \\
\texttt{\{siyulai,chenyf,jaxu\}@bjtu.edu.cn}, \\
\texttt{\{zieenyang,fandongmeng,withtomzhou\}@tencent.com} \\}

\begin{document}
\maketitle
\begin{abstract}
Word alignment which aims to extract lexicon translation equivalents between source and target sentences, serves as a fundamental tool for natural language processing. Recent studies in this area have yielded substantial improvements by generating alignments from contextualized embeddings of the pre-trained multilingual language models. However, we find that the existing approaches capture few interactions between the input sentence pairs, which degrades the word alignment quality severely, especially for the ambiguous words in the monolingual context. 
To remedy this problem, we propose \textbf{Cross-Align} to model deep interactions between the input sentence pairs, in which the source and target sentences are encoded separately with the shared self-attention modules in the shallow layers, while cross-lingual interactions are explicitly constructed by the cross-attention modules in the upper layers.
Besides, to train our model effectively, we propose a two-stage training framework, where the model is trained with a simple Translation Language Modeling (TLM) objective in the first stage and then finetuned with a self-supervised alignment objective in the second stage. 
Experiments show that the proposed Cross-Align achieves the state-of-the-art (SOTA) performance on four out of five language pairs.\footnote{The code is publicly available at: \url{https://github.com/lisasiyu/Cross-Align}}
\end{abstract}

\section{Introduction}
\begin{figure}
\centering
\includegraphics[width=0.37\textwidth]{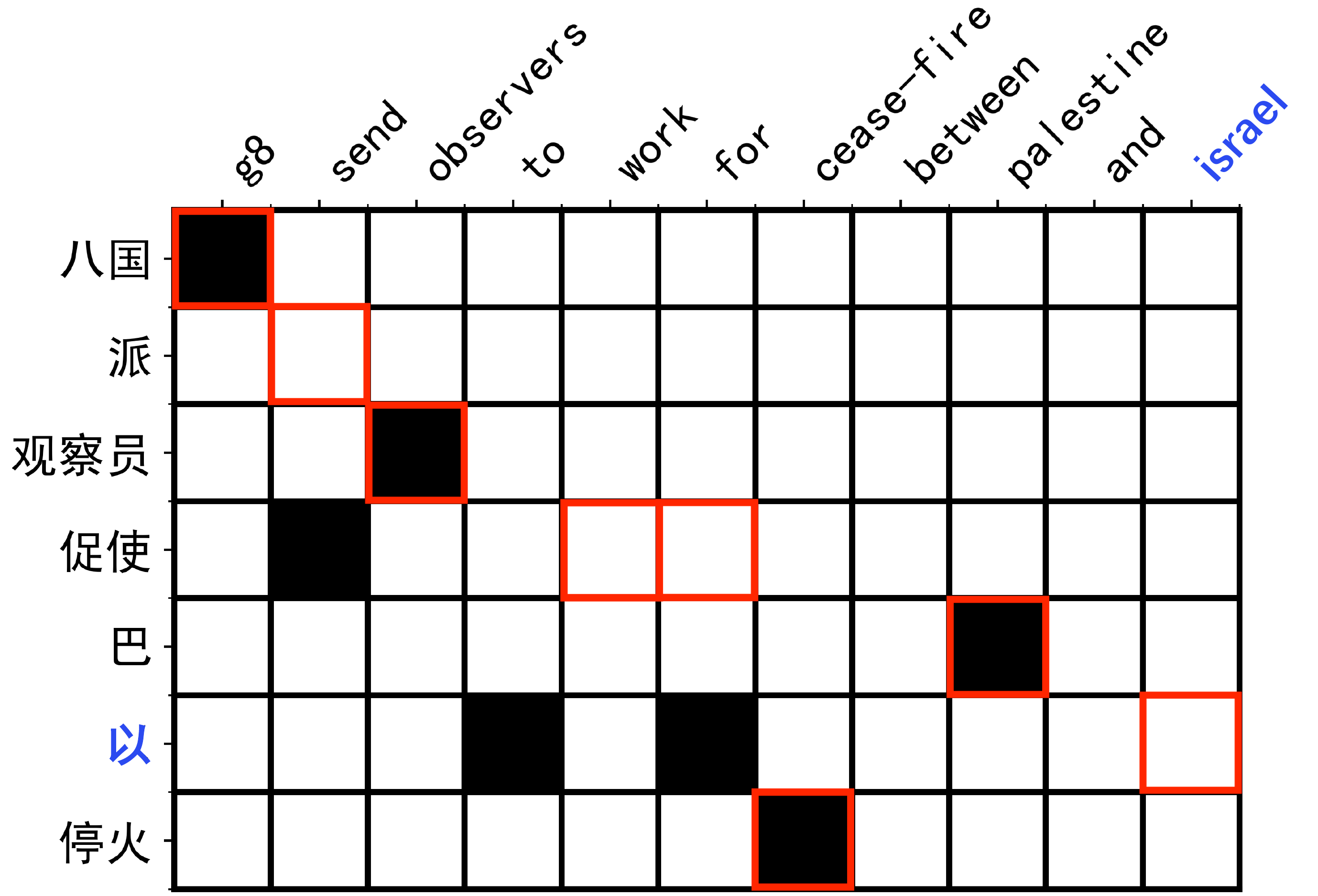}
\caption{ \label{fig:1} An example from \citet{dou2021word}. There is a misalignment between ``\begin{CJK*}{UTF8}{gbsn}{以}\end{CJK*}'' and ``to'' and ``for''. Red boxes denote the gold alignments. } 
\end{figure}

Word alignment which aims to extract the lexicon translation equivalents between the input source-target sentence pairs \citep{brown-etal-1993-mathematics,zenkel2019adding,garg2019jointly,sabet2020simalign}, has been widely used in machine translation \cite{och1999improved,arthur2016incorporating,yang-etal-2020-csp,yang2021wets}, transfer text annotations \cite{fang-cohn-2016-learning,huck-etal-2019-cross}, typological analysis \cite{lewis-xia-2008-automatically}, generating adversarial examples \cite{lai2022generating}, etc. Statistical word aligners based on the IBM translation models \cite{brown-etal-1993-mathematics}, such as GIZA++ \cite{och-ney-2003-systematic} and FastAlign \cite{dyer-etal-2013-simple}, have remained popular over the past thirty years for their good performance. Recently, with the advancement of deep neural models, neural aligners have developed rapidly and surpassed the statistical aligners on many language pairs. Typically, these neural approaches can be divided into two branches: \textbf{\emph{Neural Machine Translation (NMT) based aligners}} and \textbf{\emph{Language Model (LM) based aligners}}.

\textbf{\emph{NMT based aligners}} \cite{garg2019jointly,zenkel2020end,chen2020accurate,chen2021mask,zhang2021bidirectional} take alignments as a by-product of NMT systems by using attention weights to extract alignments. As NMT models are unidirectional, two NMT models (source-to-target and target-to-source) are required to obtain the final alignments, which makes the NMT based aligners less efficient. 
As opposed to the NMT based aligners, the \textbf{\emph{LM based aligners}} generate alignments from the contextualized embeddings of the directionless multilingual language models. They extract contextualized embeddings from LMs and induce alignments based on the matrix of embedding similarities \cite{sabet2020simalign,dou2021word}. While achieving some improvements in the alignment quality and efficiency, we find that the existing LM based aligners capture few interactions between the input source-target sentence pairs. Specifically, SimAlign \cite{sabet2020simalign} encodes the source and target sentences separately without attending to the context in the other language. \citet{dou2021word} further propose Awesome-Align, which considers the cross-lingual context by taking the concatenation of the sentence pairs as inputs during training, but still encodes them separately during inference.

However, the lack of interaction between the input source-target sentence pairs degrades the alignment quality severely, especially for the ambiguous words in the monolingual context. Figure \ref{fig:1} presents an example of our reproduced results from Awesome-Align. The ambiguous Chinese word ``\begin{CJK*}{UTF8}{gbsn}{以}\end{CJK*}'' has two different meanings: 1) a preposition (``to'', ``as'', ``for'' in English), 2) the abbreviation of the word ``\begin{CJK*}{UTF8}{gbsn}{以色列}\end{CJK*}'' (``Israel'' in English). In this example, the word  ``\begin{CJK*}{UTF8}{gbsn}{以}\end{CJK*}'' is misaligned to ``to'' and ``for'' as the model does not fully consider the word ``Israel' in the target sentence. Intuitively, the cross-lingual context is very helpful for alleviating the meaning confusion in the task of word alignment.

Based on the above observation, we propose \textbf{Cross-Align}, which fully considers the cross-lingual context by modeling deep interactions between the input sentence pairs. Specifically, Cross-Align encodes the monolingual information for source and target sentences independently with the shared self-attention modules in the shallow layers, and then explicitly models deep cross-lingual interactions with the cross-attention modules in the upper layers. Besides, to train Cross-Align effectively, we propose a two-stage training framework, where the model is trained with the simple TLM objective \cite{conneau2019cross} to learn the cross-lingual representations in the first stage, and then finetuned with a self-supervised alignment objective to bridge the gap between training and inference in the second stage. We conduct extensive experiments on five different language pairs and the results show that our approach achieves the SOTA performance on four out of five language pairs.\footnote{In Ro-En, we achieve the best performance among models in the same line, but perform a little poorer than the NMT based models which have much more parameters than ours.} Compared to the existing approaches which apply many complex training objectives, our approach is simple yet effective.

Our main contributions are summarized as follows:
\begin{itemize}[leftmargin=*]
    \itemsep=-2pt
    \item{We propose Cross-Align, a novel word alignment model which utilizes the self-attention modules to encode monolingual representations and the cross-attention modules to model cross-lingual interactions. }
    \item{We propose a two-stage training framework to boost model performance on word alignment, which is simple yet effective. }
    \item{Extensive experiments show that the proposed model achieves SOTA performance on four out of five different language pairs.}
\end{itemize}
\begin{figure*}
\centering
\includegraphics[width=\textwidth]{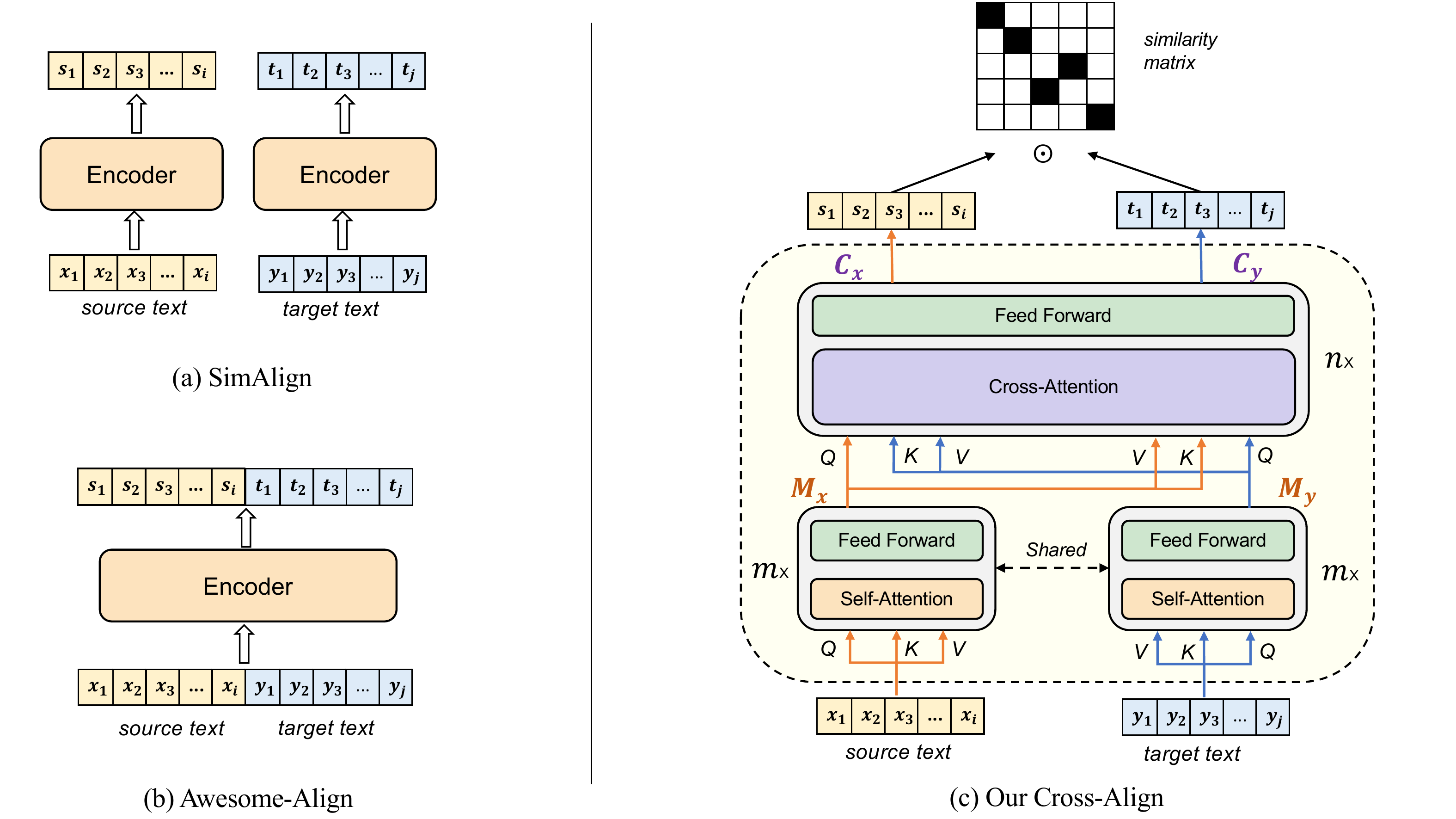}
\caption{ \label{fig:2} Comparison between different LM based aligners. (a) SimAlign \cite{sabet2020simalign} encodes source and target sentences separately. (b) AwesomeAlign \cite{dou2021word} concatenates source and target sentences together as inputs. (c) The proposed Cross-Align model. } 
\end{figure*}

\section{Related Work}
\subsection{NMT based Aligner}
Recently, there is a surge of interest in studying alignment based on the attention weights \cite{vaswani2017attention} of NMT systems. 
However, the naive attention may fails to capture clear word alignments \cite{serrano2019attention}. Therefore, \citet{zenkel2019adding} and \citet{garg2019jointly} extend the Transformer architecture with a separate alignment layer on top of the decoder, and produce competitive results compared to GIZA++.
\citet{chen2020accurate} further improve alignment quality by adapting the alignment induction with the to-be-aligned target token. Recently, \citet{chen2021mask} and \citet{zhang2021bidirectional} propose self-supervised models that take advantage of the full context on the target side, and achieve the SOTA results. Although NMT based aligners achieve promising results, there are still some disadvantages: 1) The inherent discrepancy between translation task and word alignment is not eliminated, so the reliability of the attention mechanism is still under suspicion \cite{li2019word}; 2) Since NMT models are unidirectional, it requires NMT models in both directions to obtain final alignment, which is lack of efficiency.

\subsection{LM based Aligner}
Recent pre-trained multilingual language models like mBERT \cite{devlin2018bert} and XLM-R \cite{conneau2019cross} achieve promising results on many cross-lingual transfer tasks \cite{liang2020xglue,hu2020xtreme,wang2022clidsum,wang2022survey}.
\citet{sabet2020simalign} prove that multilingual LMs are also helpful in word alignment task and  propose SimAlign to extract alignments from similarity matrices of contextualized embeddings without relying on parallel data (Figure \ref{fig:2}(a)). Awesome-Align further improves the alignment quality of LMs by crafting several training objectives based on parallel data, like masked language modeling, TLM, and parallel sentence identification task. 
Although Awesome-Align has achieved the SOTA performance among LM based aligners, we find it still has two main problems: 1) During training, they simply concatenate the source and target sentences together as the input of self-attention module (Figure \ref{fig:2}(b)). However, \citet{luo2021veco} prove that self-attention module tends to focus on their own context, while ignores the paired context, leading to few attention patterns across languages in the self-attention module. 2) During inference, they still encode the language pairs individually, which causes the cross-lingual context unavailable when generating alignments.\footnote{For Awesome-Align, concatenating the input sentence pair during inference leads to poor results compared to separately encoding. Please refer to Table \ref{tab:2} for comparison results.} Therefore, Awesome-Align models few interactions between cross-lingual pairs.
Based on the above observation, we propose Cross-Align, which aims to model deep interactions of cross-lingual pairs to solve these problems.

\section{Method}
In this section, we first introduce the model architecture and then illustrate how we extract alignments from Cross-Align. Finally, we describe the proposed two-stage training framework in detail.
\subsection{Model Architecture}

As shown in Figure \ref{fig:2}(c), Cross-Align is composed of a stack of $m$ self-attention modules and $n$ cross-attention modules \cite{vaswani2017attention}. Given a sentence ${\mathbf x}=\{x_1,x_2,\dots,x_i\}$ in the source language and its corresponding parallel sentence ${\mathbf y}=\{y_1,y_2,\dots,y_j\}$ in the target language, Cross-Align first encodes them separately with the shared self-attention modules to extract the monolingual representations, and then generate the cross-lingual representations by fusing the source and target monolingual representations with the cross-attention modules. We elaborate the self-attention module and cross-attention module as follows.

\paragraph{Self-Attention Module.}
Each self-attention module contains a self-attention sub-layer and a fully connected feed-forward network (FFN). The attention function maps a query ($\mathbf Q$) and a set of key-value ($\mathbf K$-$\mathbf V$) pairs to an output. As for self-attention, all queries, keys and values are from the same language. 
Formally, the output of a self-attention module in the $l$-th layer ($1\le l \le m$) is calculated as:
\allowdisplaybreaks
\begin{gather}
\resizebox{.9\hsize}{!}{$\mathbf {Q} = \mathbf {H}^{l-1} \mathbf {W}^Q_{s}, \mathbf {K} = \mathbf {H}^{l-1} \mathbf {W}^K_{s}, \mathbf {V} = \mathbf {H}^{l-1} \mathbf {W}^V_{s}$},\\
\resizebox{.75\hsize}{!}{$\hat{\mathbf {H}}^l= \mathrm{LN}(\mathrm{softmax}(\frac{\mathbf Q \mathbf K^T}{\sqrt{d_k}}) \mathbf{V}+\mathbf {H}^{l-1})$}, \\
\resizebox{.55\hsize}{!}{$
\mathbf {H}^l = \mathrm{LN}(\mathrm{FFN}(\hat{\mathbf {H}}^l)+\hat{\mathbf {H}}^l)$},
\end{gather}
where $\mathbf W^Q_{s},\mathbf W^K_{s},\mathbf W^V_{s}$ are parameter matrices of the self-attention module, $\mathbf H^{l-1}$ is output from previous layer, $\mathrm{LN}(\cdot)$ refers to the Layer-Normalization operation. With the above stacked $m$ self-attention modules, we get the monolingual representations $\mathbf M_x$ and $\mathbf M_y$ when $\mathbf H^0$ is set to the embeddings of $\mathbf x$ and $\mathbf y$, respectively.

\paragraph{Cross-Attention Module.}
Although the self-attention modules can effectively encode monolingual information, the interactive information between $\mathbf x$ and $\mathbf y$ is not explored. Recently, cross-attention modules have been successfully used to learn cross-modal interactions in multi-modal tasks \cite{wei2020multi,li2021align}, which motivates us to leverage cross-attention modules for exploring cross-lingual interactions in word alignment. 

Specifically, each cross-attention module contains a cross-attention sub-layer and an FFN network. Different from self-attention, the queries of cross-attention come from one language, while keys and values come from the other language. Formally, the output of a cross-attention module in the $l$-th layer ($m < l \le m+n$) is computed as:
\allowdisplaybreaks
\begin{gather}
\resizebox{.9\hsize}{!}{$\mathbf Q_{x} = \mathbf H_{x}^{l-1} \mathbf W^Q_c, 
    \mathbf K_{x} = \mathbf H_{y}^{l-1}  \mathbf W^K_c, 
    \mathbf V_{x} = \mathbf H_{y}^{l-1} \mathbf W^V_c$},\\
    \resizebox{.9\hsize}{!}{$\mathbf Q_{y} = \mathbf H_{y}^{l-1} \mathbf W^Q_c, 
    \mathbf K_{y} = \mathbf H_{x}^{l-1}  \mathbf W^K_c, 
    \mathbf V_{y} = \mathbf H_{x}^{l-1} \mathbf W^V_c$},\\
    \resizebox{.75\hsize}{!}{$
    \mathbf {\hat{H}}_{x}^l= \mathrm{LN}(\mathrm{softmax}(\frac{\mathbf Q_{x} \mathbf K_{x}^T}{\sqrt{d_k}}) \mathbf V_{x}+\mathbf H_x^{l-1}), $} \\
    \resizebox{.75\hsize}{!}{$
    \mathbf {\hat{H}}_{y}^l= \mathrm{LN}(\mathrm{softmax}(\frac{\mathbf Q_{y} \mathbf K_{y}^T}{\sqrt{d_k}}) \mathbf V_{y}+\mathbf H_y^{l-1}),  $}\\
    \resizebox{.55\hsize}{!}{$
 \mathbf H_x^l = \mathrm{LN}(\mathrm{FFN}(\mathbf {\hat{H}}_x^l)+\mathbf {\hat{H}}_x^l), $}\\
 \resizebox{.55\hsize}{!}{$
    \mathbf H_y^l = \mathrm{LN}(\mathrm{FFN}(\mathbf {\hat{H}}_y^l)+\mathbf {\hat{H}}_y^l)$},
\end{gather}
where $\mathbf W^Q_c,\mathbf W^K_c,\mathbf W^V_c$ are parameter matrices of the cross-attention module, $\mathbf H_x^{l-1}$ is output from the previous layer corresponding to ${\mathbf x}$ and $\mathbf H_y^{l-1}$ is output from the previous layer corresponding to ${\mathbf y}$. With the above stacked $n$ cross-attention modules, we get the cross-lingual representation $\mathbf C_x$ and $\mathbf C_y$ by setting $\mathbf H_x^m$ and $\mathbf H_y^m$ to  $\mathbf M_x$ and $\mathbf M_y$, respectively. 

\subsection{Alignments Extraction} \label{sec:3.2}
The proposed Cross-Align aims to extract alignments from the input sentence pair $\mathbf x$ and $\mathbf y$, and we illustrate the extraction method as follows.

Firstly, we extract the hidden states $\mathbf s=\{s_1,s_2,\dots,s_i\}$ and $\mathbf t=\{t_1,t_2,\dots,t_j\}$ for ${\mathbf x}$ and ${\mathbf y}$ respectively. Secondly, we get a similarity matrix $S_{I\times J}$ by computing the dot products between $\mathbf {s}$ and $\mathbf {t}$ and apply the $softmax$ normalization to convert $S_{I\times J}$ into the source-to-target probability matrices $P^f_{I \times J}$ and target-to-source probability matrices $P^b_{I \times J}$. After that, we obtain the final alignment matrix $G_{I \times J}$ by taking the intersection of the two matrices following \citet{dou2021word}:
\begin{equation}
    G = (P^f > \tau) * (P^b > \tau),
\end{equation}
where $\tau$ is a threshold. $G_{ij}=1$ means $x_i$ and $y_j$ are aligned. Note that the current alignments generated from Cross-Align are on BPE-level. We follow previous work to convert BPE-level alignments to word-level alignments \cite{dou2021word,zhang2021bidirectional} by adding an alignment between a source word and a target word if any parts of these two words are aligned. 
\subsection{Two-stage Training Framework}
This sub-section describes the proposed two-stage training framework. In the first stage, the model is trained with TLM to learn the cross-lingual representations. After the first training stage, the model is then finetuned with a self-supervised alignment objective to bridge the gap between the training and inference.

\paragraph{Stage1: Translation Language Modeling.} 
TLM is a simple training objective first proposed by \citet{conneau2019cross} for learning cross-lingual representations of LMs.
Since Cross-Align aims to learn interactions between the input sentence pairs, TLM is a suitable objective for effectively training Cross-Align. Different from \citet{conneau2019cross} which train TLM objective based on the self-attention modules, Cross-Align applies the cross-attention modules to enforce the model to infer the masked tokens based on the cross-lingual representations $\mathbf C_x$ and $\mathbf C_y$ , encouraging deep interactions between the input sentence pair. 

Following the previous works \cite{devlin2018bert,conneau2019cross}, we choose 15\% of the token positions randomly for both $\mathbf{x}$ and $\mathbf{y}$. For each chosen token, we replace it with the \texttt{[MASK]} token 80\% of the time, a random token 10\% of the time, and remain token 10\% of the time. The model is trained to predict the original masked words based on the bilingual context. Thus, the training objective can be formulated as:
\allowdisplaybreaks
\begin{equation} 
\begin{split}
        \mathcal{L}_{TLM} = &-{\rm log}P({\mathbf x}|\hat{\mathbf x}, \hat{\mathbf y}; \mathbf{\theta_s},\mathbf{\theta_c}) \\
        &-{\rm log}P({\mathbf y}|\hat{\mathbf x}, \hat{\mathbf y}; \mathbf{\theta_s},\mathbf{\theta_c}),
\end{split}
\end{equation}
where $\hat{\mathbf x}$ and $\hat{\mathbf y}$ are the masked sentences for ${\mathbf x}$ and ${\mathbf y}$ respectively, $\mathbf{\theta_s}$ denotes all the parameters of the $m$ self-attention modules, and $\mathbf{\theta_c}$ represents the parameters of $n$ cross-attention modules.
\paragraph{Stage2: Self-Supervised Alignment.}
In the first training stage, the model is trained with TLM by feeding the masked sentence pairs as input. However, the model is required to extract the alignments from the original sentence pairs during inference. Therefore, there is a gap between the training and inference which may limit the alignment quality. To solve this problem, we propose a self-supervised alignment (SSA) objective in the second stage. SSA takes the alignments generated by the model trained in the first stage as golden labels and trains the model with the alignment task directly in this stage. 

As previous studies \cite{sabet2020simalign,dou2021word} have shown that the middle layer of LM always has better alignment performance than the last layer, we take the $c$-th layer of Cross-Align as the alignment layer to train the alignment objective, where $c$ ($1 \leq c \leq m+n$) is a hyper-parameter chosen from the experimental results.\footnote{The analysis about the alignment layer is conducted in Section \ref{sec:5.2}.} From the alignment layer of the first-stage model, we extract the 0-1 alignment labels $G_{I\times J}$ 
according to extraction method described in Section \ref{sec:3.2}.\footnote{Now the alignment labels are on word level, while SSA objective is on BPE level, so we convert labels back to BPE-level as follows: a source BPE token is aligned to a target BPE token if their corresponding source word and target word are aligned. Besides, a target BPE tokens will be aligned with [CLS] token, if its corresponding target word is not aligned with any source word.} 
$P^f_{I\times J}$ and $P^b_{I\times J}$ denotes the source-to-target and target-to-source probability matrices extracted from the alignment layer of the current model, respectively.
Following \citet{garg2019jointly}, we optimize the alignment objective by minimizing the cross-entropy loss: 
\allowdisplaybreaks
\begin{equation}
\begin{split}
    \mathcal{L}_{SSA} = &-\frac{1}{I}\sum_{i=1}^I\sum_{j=1}^JG_{ij}{\rm log}(P^f_{ij}) \\
    &-\frac{1}{J}\sum_{j=1}^J\sum_{i=1}^IG_{ij}{\rm log}(P^b_{ij}).
\end{split}
\end{equation}
To alleviate the catastrophe of forgetting knowledge learned by the TLM, we only finetune the alignment layer and freeze other layers of the model. 
With the SSA objective, Cross-Align directly learns the word alignment task in the alignment layer instead of the masked language modeling, making the training consistent with the inference process. During inference, we extract hidden states from the alignment layer and get the final alignments.

\section{Experimental Settings}
In this section, we first describe the details of datasets and implementation, then present the baselines, and finally introduce evaluation measures.

\subsection{Datasets}
We conduct our experiments on five publicly available datasets,  including German-English (De-En), English-French (En-Fr), Romanian-English (Ro-En), Chinese-English (Zh-En), and Japanese-English (Ja-En). The training sets only contain the parallel sentences without word alignment labels, the development and test sets contain parallel sentences with gold word alignment labels annotated by experts. Table \ref{tab:1} gives the detailed data statistics. Considering that De-En, En-Fr, and Ro-En do not have development sets, we use Zh-En development sets to tune the hyper-parameters for them.

\begin{table}
\centering
\scalebox{0.85}{
\begin{tabular}{l c c c c c}
    \toprule[1.2pt]
    {\textbf{Dataset}} & \textbf{De-En} & \textbf{En-Fr} & \textbf{Ro-En}  &\textbf{Zh-En} & \textbf{Ja-En} \\
    \midrule
    training & 1.9M & 1.1M & 447k & 1.2M & 442k \\
    dev & - & - & -  & 450 & 653 \\
    test & 508 & 447 & 248 &  450 & 582\\
    \bottomrule[1.2pt]
    \end{tabular}
}
\caption{\label{tab:1} The number of sentences in each dataset. }
\end{table}
\begin{table*} 
\centering
\begin{tabular}{l c c c c c}
    \toprule[1.2pt]
    {\textbf{Method}} & \textbf{De-En} & \textbf{En-Fr} & \textbf{Ro-En} & \textbf{Zh-En} & \textbf{Ja-En}\\
    \midrule
   \emph{Statistic Based}\\
    FastAlign \cite{dyer-etal-2013-simple} & ~~26.2$^{\dag}$ & ~~10.5$^{\dag}$ & ~~31.4$^{\dag}$ & ~~23.7$^{\dag}$ & ~~51.1$^{\dag}$\\
    GIZA++ \cite{och-ney-2003-systematic} & ~~18.9$^{\dag}$ & ~~5.5$^{\dag}$ & ~~26.6$^{\dag}$ & ~~19.4$^{\dag}$ & ~~48.0$^{\dag}$\\
    \midrule
    \emph{Neural Machine Translation Based}\\
    \citet{zenkel2019adding} & 21.2 & 10.0 & 27.6 & - & -\\
    \citet{garg2019jointly} & 20.2 & 7.7 & 26.0 & ~~22.5$^{\dag}$ & ~~49.8$^{\dag}$\\
    \citet{zenkel2020end} & 16.3 & 5.0 & 23.4 & - & -\\
    SHIFT-AET \cite{chen2020accurate} & 15.4 & 4.7 & 21.2 & ~~18.6$^{\dag}$ & ~~44.3$^{\dag}$ \\
    \citet{zhang2021bidirectional} & 14.3 & 6.7 & \textbf{18.5} & - & -\\
    MASK-ALIGN \cite{chen2021mask} & 14.4 & 4.4 &19.5 & 13.8 & ~~43.5$^{\dag}$\\
 \midrule
\emph{Multilingual Language Model Based}\\
    SimAlign \cite{sabet2020simalign} & 18.8 & 7.6 & 27.2 & ~~21.6$^{\dag}$ & 46.6\\
    Awesome-Align \cite{dou2021word} & 15.6 & 4.4 & 23.0 & ~~12.9$^{\dag}$ & 38.4\\
    Awesome-Align (\emph{concatenation}) & ~~16.8$^{\dag}$ &  ~~4.7$^{\dag}$ &  ~~23.2$^{\dag}$  & ~~14.2$^{\dag}$ & ~~39.3$^{\dag}$\\
    \textbf{Cross-Align (ours)} & \textbf{13.6} & \textbf{3.4} & 20.9 & \textbf{10.1} & \textbf{35.4}\\
    \bottomrule[1.2pt]
    \end{tabular}
\caption{\label{tab:2} AER on the test sets with different alignment methods. The lower AER, the better performance. We highlight the best results for each language pair in \textbf{bold}. To make a fair comparison, we only report the results for all baselines under bilingual settings and without guidance from external word alignment tools. `Awesome-Align  (\emph{concatenation})' means the source and target sentences are concatenated as inputs during inference. `$^\dag$' denotes the re-implement results based on their released code for those results not reported in the original paper.   }
\end{table*}
\subsection{Implementation Details}
Our implementation is based on the code base released by Awesome-Align.\footnote{\url{https://github.com/neulab/awesome-align}} We randomly initialize the parameters of cross-attention modules and leverage the pre-trained mBERT-base \cite{devlin2018bert} to initialize the rest parameters of our Cross-Align. The AdamW \cite{kingma2014adam} is used as the optimizer, and the learning rate is set to 5e-4 and 1e-5 for the two stages of training, respectively. The batch size per GPU is set to 12 and gradient accumulation steps is set to 4. All models are trained on 8 NVIDIA Tesla V100 (32GB) GPUs. We train 2 epochs for each language pair in the first stage and then finetune 1 epoch in the second stage. The number of self-attention layers $m$ and cross-attention layers $n$ are set to 10 and 2, respectively. The alignment layer $c$ is set to 11. In the first stage, the threshold of extraction $\tau$ is set to 0.001. In the second stage, $\tau$ is set to 0.15. 
\subsection{Baselines}
To test the effectiveness of Cross-Align, we take the current three types of aligners as baselines.
\paragraph{Statistic based methods:}
\begin{itemize}[leftmargin=*,topsep = 1 pt]
    \itemsep=-2pt
    \item{FastAlign \cite{dyer-etal-2013-simple} and GIZA++ \cite{och-ney-2003-systematic} are two popular statistical aligners that are implementations of IBM model.}
\end{itemize}
\paragraph{NMT based methods:}
\begin{itemize}[leftmargin=*,topsep = 1 pt]
\itemsep=-2pt
\item{\citet {zenkel2019adding} and \citet{zenkel2020end} propose to add an extra attention layer on top of NMT model which produces translations and alignment simultaneously.}
    \item{\citet{garg2019jointly} propose a multi-task framework to align and translate with transformer models jointly.}
    \item{SHIFT-AET: \citet{chen2020accurate}} induce alignments when the to-be-aligned target token is the decoder input instead of the output. 
    \item{\citet{zhang2021bidirectional} predict the target word based on the source and both left-side and right-side target context to produce attention. }
    \item{MASK-ALIGN: \citet{chen2021mask} proposed a self-supervised word alignment model that takes advantage of the full context on the target side.}
\end{itemize}
\paragraph{LM based methods:}
\begin{itemize}[leftmargin=*,topsep = 1 pt]
\itemsep=-2pt
    \item{SimAlign: \citet{sabet2020simalign} extract alignment from multilingual pre-trained language models without using parallel training data. }
    \item{Awesome-Align: \citet{dou2021word} further finetune multilingual pre-trained language models on parallel corpora to get better alignments}
\end{itemize}
\subsection{Evaluation Measures}
Alignment Error Rate (AER) is the standard evaluation measure for word alignment \cite{och-ney-2003-systematic}. The quality of an alignment $A$ is computed by:
\begin{equation}
    \text {AER} = 1 - \frac{|A \cap S| + |A \cap P|}{|A|+ |S|},
\end{equation}
where $S$ (sure) are unambiguous gold alignments and $P$ (possible) are ambiguous gold alignments.

\begin{table}
\centering
\scalebox{0.85}{
\begin{tabular}{l c c c c c}
    \toprule[1.2pt]
    {\textbf{Objective}} & \textbf{De-En} & \textbf{En-Fr} & \textbf{Ro-En} & \textbf{Zh-En} &\textbf{Ja-En} \\
    \midrule
    None & 59.5 & 38.1 & 82.1 & 73.6 & 83.0\\
    +TLM & 15.3 & 4.3 & 22.1 & 12.5 & 37.1\\
    ++SSA & 13.6  & 3.4  & 20.9 & 10.1 & 35.4\\
    \bottomrule[1.2pt]
    \end{tabular}
}
\caption{\label{tab:3} Ablation studies on the two-stage training objective. `None' means the naive Cross-Align without further training on parallel corpus.  `+TLM' means training Cross-Align on TLM objective. `++SSA' denotes further finetuned on SSA objective after TLM.}
\end{table}
\section{Results and Analysis}
\subsection{Main Results}

Table \ref{tab:2} compares the performance of Cross-Align against statistical aligners, NMT based aligners, and LM based aligners. 
We can see that Cross-Align significantly outperforms the statistical method GIZA++ by 2.1\textasciitilde12.6 AER points across different language pairs. 
Compared to other LM based aligners, Cross-Align also achieves substantial improvement on all datasets. For example, on the Ja-En dataset, Cross-Align achieves 3.0 AER points improvement compared to Awesome-Align, demonstrating that modeling cross-lingual interactions based on the bilingual context is crucial for improving alignment quality. 
Compared to the strong NMT baselines with more parameters, we find Cross-Align still achieves the best results on all language pairs except Ro-En. We suppose the reason is that the parameters of cross-attention modules are initialized randomly, and the data size of Ro-En is too small to sufficiently train these parameters, resulting in unsatisfactory results compared to NMT based methods. We tried to use the self-attention parameters of mBERT to initialize it, but the results are not as good as random initialization. We will investigate the word alignment on low-resource language pairs in future work.

\subsection{Analysis} \label{sec:5.2}
\paragraph{Ablation Study.}
To understand the importance of the two-stage training objective, we conduct an ablation study by training multiple versions of the alignment models with some training stages removed. For all models, we extract the alignments on the alignment layer. The experimental results are shown in Table \ref{tab:3}. From Table \ref{tab:3}, we can find that the naive Cross-Align without training on the parallel corpus achieves very bad performance (see line ``None'' in Table \ref{tab:3}). This is mainly because that the cross-attention modules are initialized randomly. TLM objective plays a critical role in training Cross-Align since it greatly improves the quality of alignment across all language pairs (see line``+TLM'' in Table \ref{tab:3}). 
In the second stage, the SSA objective further improves the performance by 0.9\textasciitilde2.4 AER points (see line ``++SSA''). This shows that bridging the gap between the training and inference is helpful to the final alignment performance.

\begin{figure}
\centering
\includegraphics[width=0.45\textwidth]{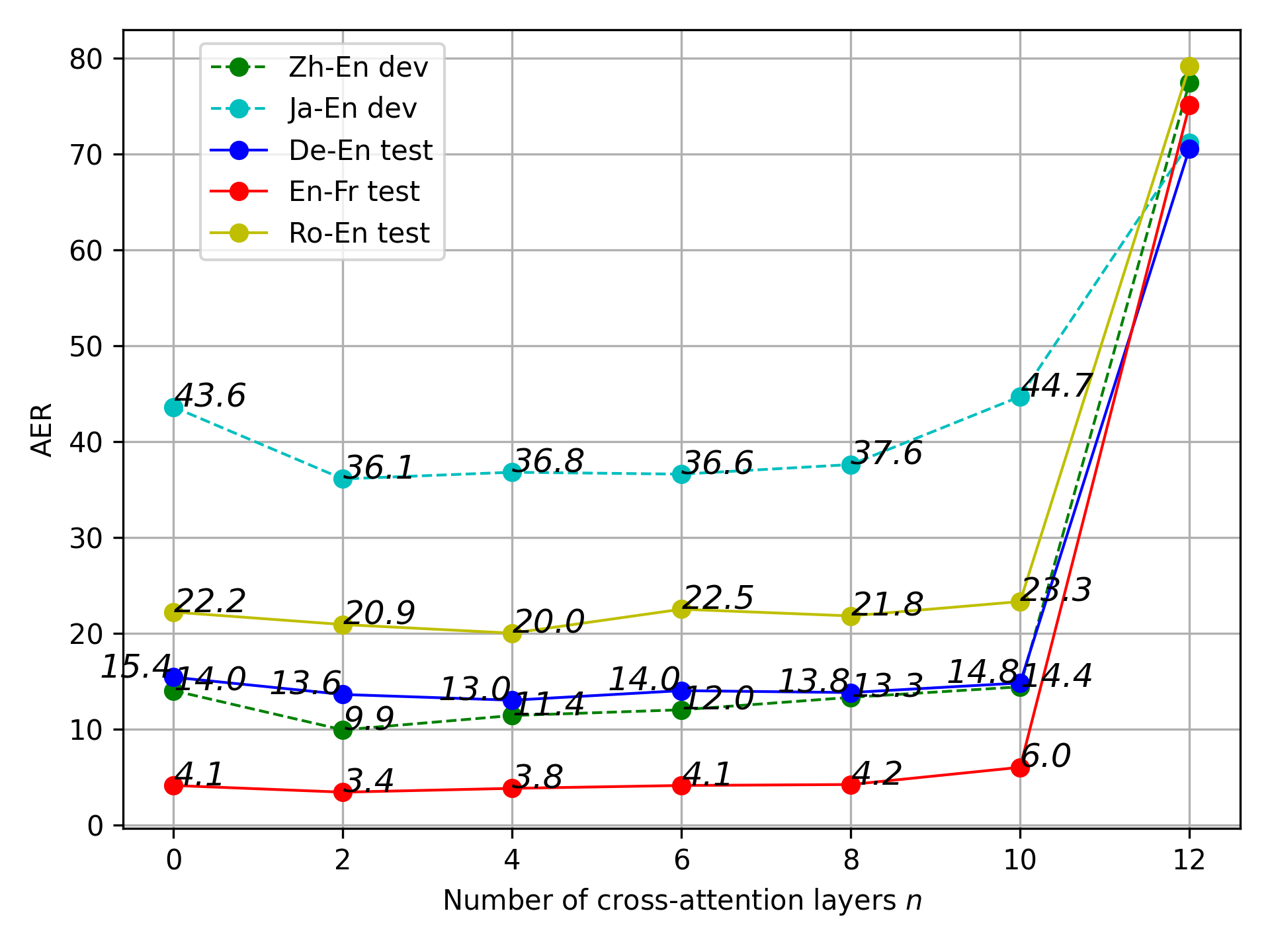}
\caption{\label{fig:4} Word alignment performance with different number of cross-attention layers $n$.} 
\end{figure}

\begin{figure*}
\centering
\includegraphics[width=\textwidth]{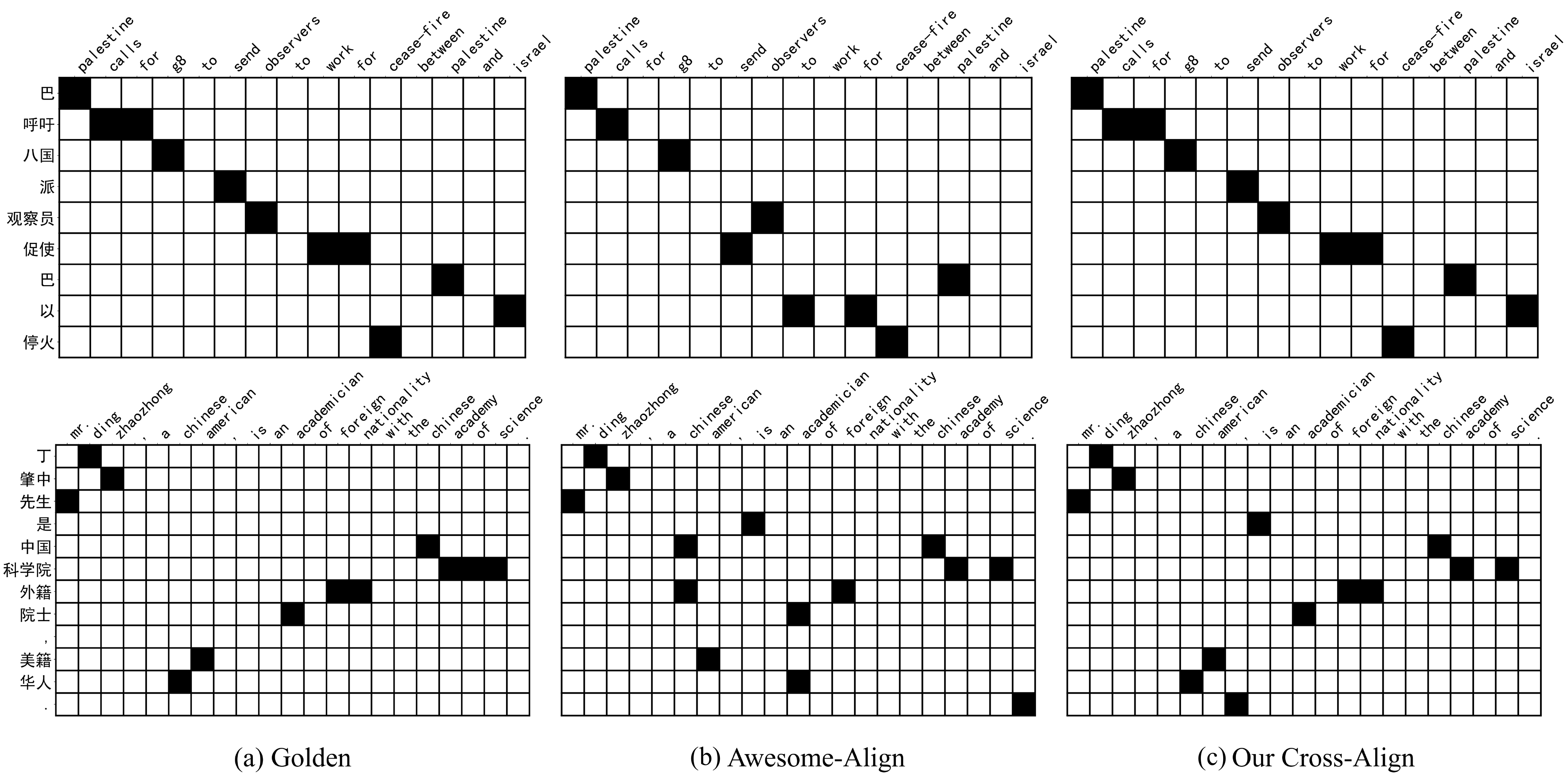}
\caption{\label{fig:6} Two examples from Zh-En alignment test set. (a) Gold alignments. (b) Results of Awesome-Align. (c) Results of Cross-Align. } 
\end{figure*}

\paragraph{Number of Cross-Attention Layers.}
Since the self-attention and cross-attention modules play different roles in the final alignments, we are curious about how the number of cross-attention layers affects the final alignment performance. We investigate this problem by studying the alignment performance with different $n$, where $n$ ranges from 0 to 12 with an interval of 2. Meanwhile, we keep $m+n=12$ to ensure that Cross-Align has the same number of layers as mBERT. Figure \ref{fig:4} shows the AER results on the dev sets with different $n$. For a more comprehensive analysis, we also show the results on the test sets for language pairs without dev sets. As shown in Figure \ref{fig:4}, Cross-Align degenerates into the separate encoding framework when $n=0$, achieving bad alignment performance. This shows that modeling the cross-lingual interactions is very helpful for enhancing the alignment performance. Additionally, when $n=12$, the performance drops sharply, which shows that the monolingual representations built by the self-attention modules are necessary for the following cross-attention modules to generate reliable cross-lingual representations. Almost all of the language pairs achieve the best performance when $n$ is set around 2 and there is a trade-off between the self-attention and cross-attention module layers.

\begin{figure}
\centering
\includegraphics[width=0.45\textwidth]{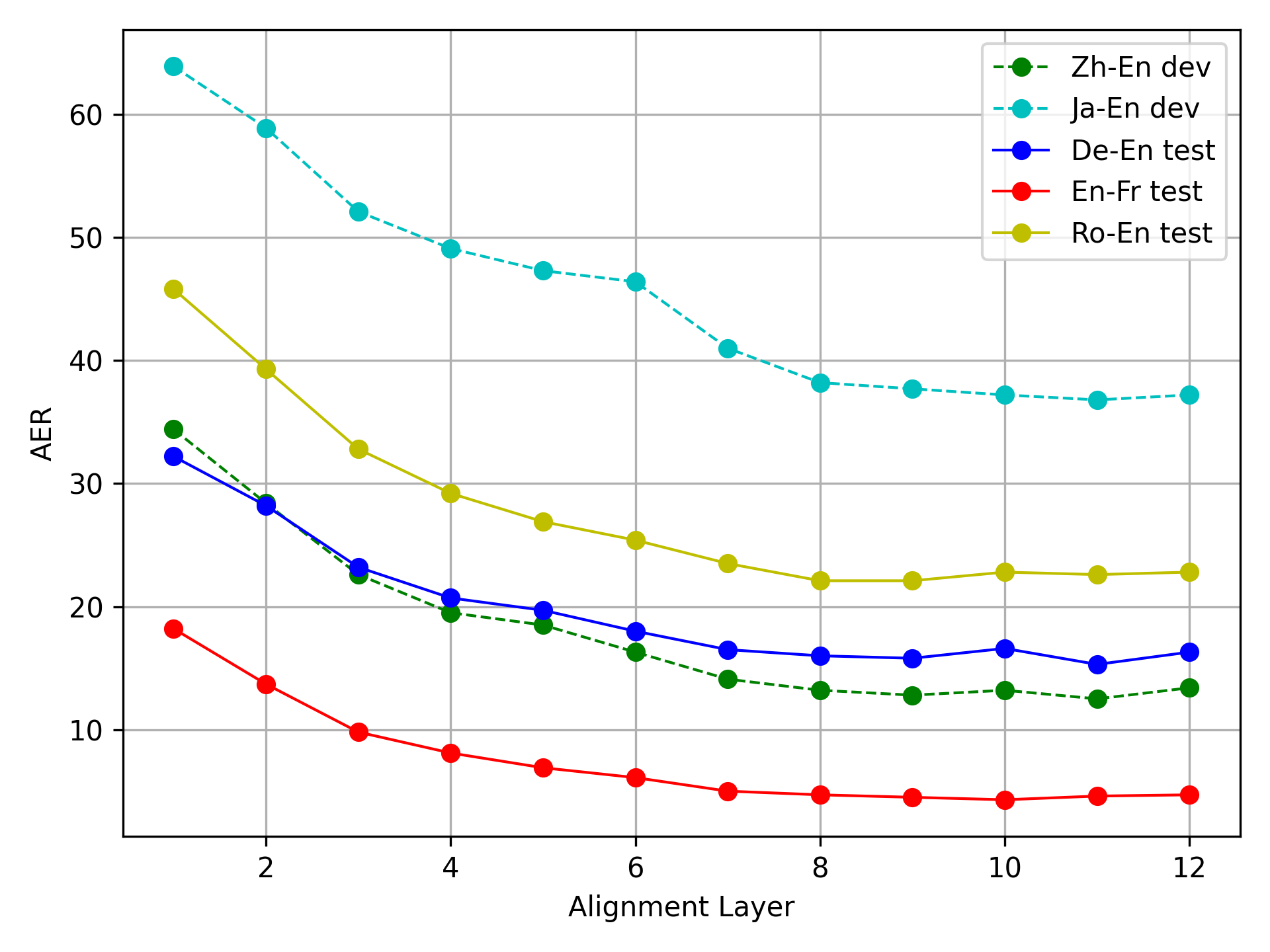}
\caption{\label{fig:5} Word alignment performance across different alignment layers of Cross-Align in the first stage.} 
\end{figure}

\paragraph{Alignment Layer.}
After the training of TLM, we need to decide the alignment layer $c$ used to generate alignments. Figure \ref{fig:5} shows the AER results with $c$ varying from 0 to 12. We observe that Cross-Align obtains the best performance when $c$ is set around 11. This observation is consistent with previous studies \cite{sabet2020simalign,conneau-etal-2020-emerging}. For Cross-align, the context representations in the lower self-attention layers are too language-specific to achieve high-quality alignment performance. In the upper cross-attention layers, the contextual representations are too specialized in the masked language modeling. The contextualized representations in the middle of cross-attention layers contain rich cross-lingual knowledge that help generate high-quality alignments.

\subsection{Case Study}
In Figure \ref{fig:6}, we present two examples from different alignment methods on Zh-En test set. In the first example, Cross-Align correctly aligns the ambiguous Chinese word ``\begin{CJK*}{UTF8}{gbsn}{以}\end{CJK*}'' to ``Israel'' and ``\begin{CJK*}{UTF8}{gbsn}{促使}\end{CJK*}'' to ``work for'' based on the bilingual context, but Awesome-Align does not. In the second example, there are two ``chinese'' in the target sentence with different meanings. Due to lack of cross-lingual context, Awesome-Align could not distinguish the difference and wrongly aligns ``\begin{CJK*}{UTF8}{gbsn}{中国}\end{CJK*}'' to both ``chinese'', but Cross-Align gives correct alignments for them. It demonstrates that learning interaction knowledge between the source-target sentence pairs is beneficial to word alignment.
\section{Conclusion}
This paper presents a novel LM based aligner named Cross-Align, which models deep interactions between the input sentence pairs. Cross-Align first encodes the source and target sentences separately with the shared self-attention modules in the shallow layers, then explicitly constructs cross-lingual interactions with the cross-attention modules in the upper layers.
Additionally, we propose a simple yet effective two-stage training framework, where the model is first trained with a simple TLM objective and then finetuned with a self-supervised alignment objective.
Experimental results show that Cross-Align achieves new SOTA results on four out of language pairs.
In future work, we plan to improve the alignment quality on more low-resource language pairs.
\section*{Limitations}
Although the proposed Cross-Align has achieved promising results, we find it still has two main limitations. Firstly, Cross-Align has limited performance in low-resource language pairs like Ro-En and Ja-En, as shown in Table \ref{tab:2}. We hypothesize the reason is that the cross-attention modules of Cross-Align are randomly initialized, so it needs a large number of data to train. We tried to use the self-attention parameters of mBERT to initialize it, but the results are not as good as random initialization. Secondly, we find current LM based aligners including Cross-Align have bad performance for phrase alignments. As shown in the second example in Figure \ref{fig:6}, ``academy of science'' is a phrase that should be aligned to the Chinese word ``\begin{CJK*}{UTF8}{gbsn}{科学院}\end{CJK*}'', but Cross-Align only aligns part of it. It is because Cross-Align generates subword-level alignments without considering the word-level and phrase-level information. In future work, we will investigate these two limitations and further improve the quality of alignments.

\section*{Acknowledgements}
The research work descried in this paper has been supported by the National Key R\&D Program of China (2020AAA0108001) and the National Nature Science Foundation of China (No. 61976016, 61976015, and 61876198). The authors would like to thank the anonymous reviewers for their valuable comments and suggestions to improve this paper.


\bibliography{anthology,custom}
\bibliographystyle{acl_natbib}




\end{document}